\pdfoutput=1

\documentclass[11pt]{article}

\usepackage[final]{acl}

\usepackage{times}
\usepackage{latexsym}

\usepackage[T1]{fontenc}

\usepackage[utf8]{inputenc}
\usepackage{amsmath}
\usepackage{amsthm}
\usepackage{amssymb}
\usepackage{microtype}

\usepackage{inconsolata}

\usepackage{graphicx}

\usepackage{booktabs}
\usepackage{multirow}
\usepackage{caption}
\usepackage{float}
\usepackage{array} 

\usepackage{soul}
\usepackage{arydshln}
\usepackage{rotating} 
\usepackage{dblfloatfix}

\usepackage{hyperref}

\title{Rewarding What Matters: Step-by-Step Reinforcement Learning for Task-Oriented Dialogue}

\author{Huifang Du\textsuperscript{*} \\
  Tongji University \\
  \texttt{duhuifang@tongji.edu.cn} \\\And
  Shuqin Li\textsuperscript{*} \\
  Hangzhou Dianzi University \\
  \texttt{shuqinlee9683@gmail.com} \\\And
  Minghao Wu \\
  Monash University \\
  \texttt{minghao.wu@monash.edu} \\\AND
  Xuejing Feng \\
  Tongji University\\
  \texttt{fengxuejing@tongji.edu.cn} \\\And
  Yuan-Fang Li \\
  Monash University \\
  \texttt{yuanfang.li@monash.edu} \\\And
  Haofen Wang \\
  Tongji University \\
  \texttt{carter.whfcarter@gmail.com} 
}

\begin{document}

\renewcommand{\tableautorefname}{Table}
\renewcommand{\sectionautorefname}{Section}
\renewcommand{\subsectionautorefname}{Section}
\renewcommand{\subsubsectionautorefname}{Section}
\renewcommand{\figureautorefname}{Figure}
\renewcommand{\equationautorefname}{Equation}

\maketitle
\begingroup
\renewcommand\thefootnote{\fnsymbol{footnote}}
\footnotetext[1]{These authors contributed equally to this work.}
\endgroup

\begin{abstract}
Reinforcement learning (RL) is a powerful approach to enhance task-oriented dialogue (TOD) systems. However, existing RL methods tend to mainly focus on generation tasks, such as dialogue policy learning (DPL) or response generation (RG), while neglecting dialogue state tracking (DST) for understanding. This narrow focus limits the systems to achieve globally optimal performance by overlooking the interdependence between understanding and generation. Additionally, RL methods face challenges with sparse and delayed rewards, which complicates training and optimization. To address these issues, we extend RL into both understanding and generation tasks by introducing step-by-step rewards throughout the token generation. The understanding reward increases as more slots are correctly filled in DST, while the generation reward grows with the accurate inclusion of user requests. Our approach provides a balanced optimization aligned with task completion. Experimental results demonstrate that our approach effectively enhances the performance of TOD systems and achieves new state-of-the-art results on three widely used datasets, including MultiWOZ2.0, MultiWOZ2.1, and In-Car. Our approach also shows superior few-shot ability in low-resource settings compared to current models.


\end{abstract}

\begin{figure}[t]
  \includegraphics[width=\columnwidth]{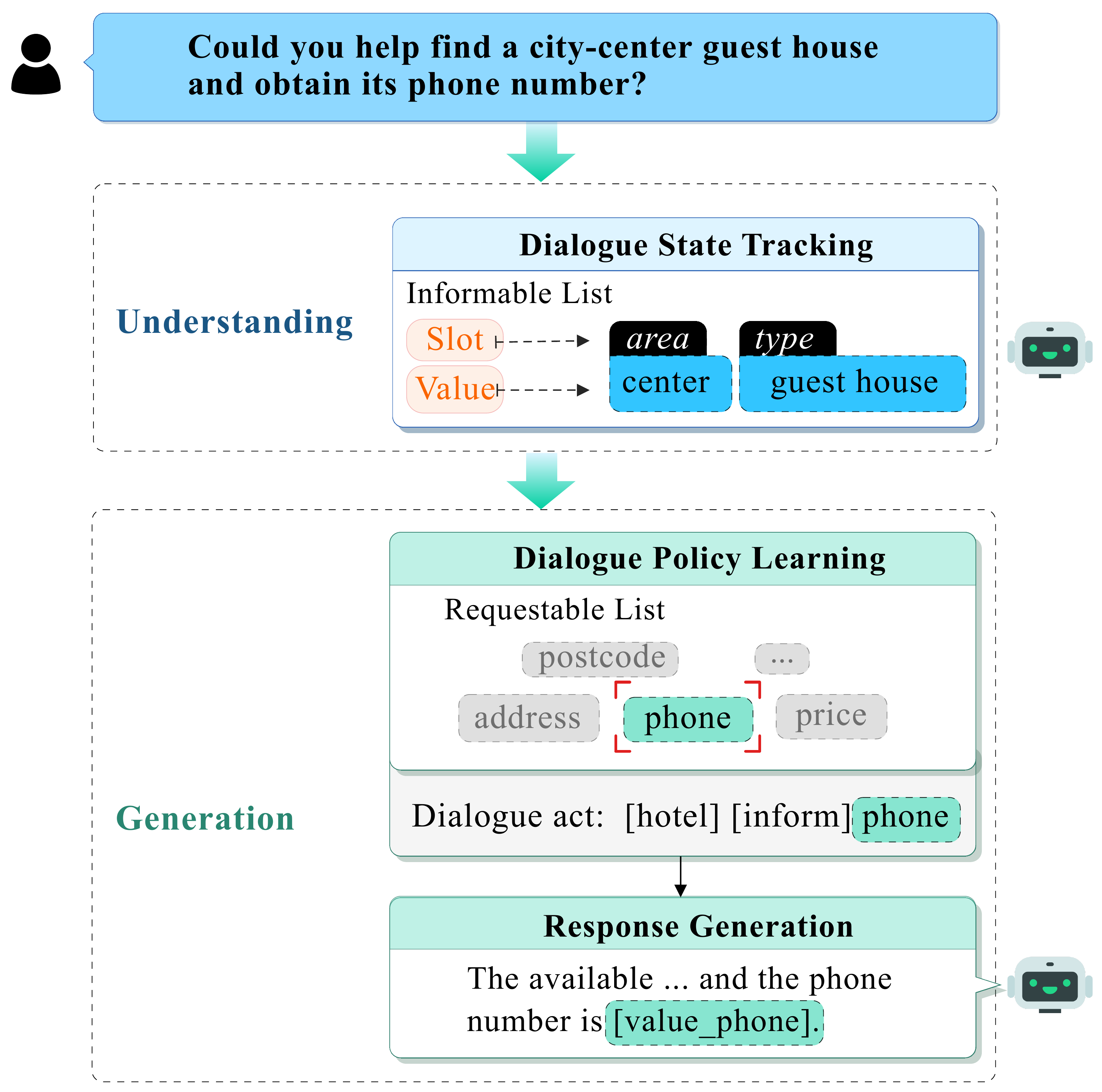}
  \caption{A task-oriented dialogue system needs to successfully perform both understanding and generation to achieve its dialogue goals.}
  \label{fig:teaser}
\end{figure}

\section{Introduction}
The rapid advancements in pre-trained language models (PLMs) have significantly influenced a variety of real-world applications \cite{devlin2018bert, raffel2020exploring, chung2024scaling}. Among these, the development of task-oriented dialogue (TOD) systems stands out as particularly impactful \cite{wen2017network, hosseini2020simple}. Typically, a TOD system comprises several components \cite{he2022galaxy, feng2023fantastic} as shown in \autoref{fig:teaser}, including dialogue state tracking (DST) for understanding user’s belief state \cite{chen2020schema, guo2023learning}, dialogue policy learning (DPL) for generating dialogue acts \cite{zhao2024bootstrapped, zhang2019budgeted}, and response generation (RG) for generating system responses \cite{pei2020retrospective, chen2019working}. More recently, there has been growing interest in constructing end-to-end (E2E) TOD systems based on PLMs to equip models with all these essential capabilities \cite{he2022galaxy, hosseini2020simple, feng2023fantastic, yu2023krls}.

Building on the advancements in TOD systems discussed earlier, recent research explores the use of offline reinforcement learning (RL) to optimize TOD systems further learning goal-oriented conversational strategies \cite{lu2019goal,jang2021gpt,feng2023fantastic}. However, current RL approaches typically focus on enhancing the generation component, such as generating dialog acts (DPL task)  \cite{NEURIPS2023_c5601d99} or system response (RG task) \cite{yu2023krls}. This biased focus prevents the systems from reaching optimal performance by ignoring the crucial interdependence between understanding and generation.
Furthermore, RL for TOD systems often faces issues with sparse and delayed rewards \cite{lu2019goal, DBLP:journals/corr/abs-2311-18232}, which are only provided upon reaching the goal at the dialogue or turn level \cite{kwan2023survey, lu2019goal, DBLP:journals/corr/abs-2311-18232}. This leads to insufficient exploration and unstable training for RL. While many efforts have tried to mitigate these reward issues to offer dense rewards, the design of the reward function in these methods tends to be complex, which may limit the method's generalization \cite{li2020rethinking,feng2023fantastic}.


In this work, we propose to design a simple but effective reward function to jointly optimize both \textit{understanding} and \textit{generation} components in an \textit{end-to-end} manner to achieve the globally optimal performance. We propose the combination of understanding reward and generation reward throughout per token generation to reinforce the learning step by step. The understanding reward is the growing proportion of correctly filled slots in the DST process, while the generation reward is measured by the correct inclusion of the user requests in the DPL and RG process.  We conduct extensive experiments using two model backbones, the Flan-T5 base and Flan-T5 large models \cite{chung2024scaling}, on three widely used benchmarks: MultiWOZ2.0, MultiWOZ2.1, and In-Car. The results show that our approach significantly improves model performance against strong baselines, establishing new state-of-the-art results. We also show that our approach outperforms current models in low-resource conditions, highlighting its adaptability in real-world scenarios where data is limited.


Our contributions to this work are summarized as follows:
\begin{itemize}
 \item We introduce a novel approach that integrates RL into both understanding (DST) and generation (DPL and RG) components in an end-to-end manner, which promotes a balanced optimization for TOD systems.
 
 \item To tackle the challenges of sparse and delayed rewards in RL for TOD systems, we propose a combined reward mechanism that provides progressive feedback during token generation. This step-by-step reward significantly enhances efficiency.

 \item Experimental results show that our approach establishes new state-of-the-art results on multiple benchmarks (MultiWOZ2.0, MultiWOZ2.1, and In-Car). Furthermore, the method shows superior performance in low-resource conditions.
 
\end{itemize}




\section{Related Work}
In this section, we review works on TOD systems utilizing both pipeline and E2E methods, the integration of reinforcement learning (RL), and the design of reward functions for RL. Additionally, we discuss the role of large language models (LLMs) in TOD systems.

\paragraph{Pipeline and End-to-End Approaches.} Pipeline approaches are characterized by their modular structure, where dialogue state tracking (DST) \cite{chen2020schema,guo2023learning}, dialogue policy learning (DPL) \cite{zhao2024bootstrapped,zhang2019budgeted}, and response generation (RG) \cite{pei2020retrospective,chen2019working} are processed sequentially. They offer interpretability and modularity but often struggle to capture the overall context of conversations \cite{kwan2023survey}. In contrast, E2E approaches directly map input utterances to system responses without explicit intermediate representations \cite{he2022galaxy,yang2021ubar,he2022unified}. Some models, such as SPACE-3 \cite{he2022unified}, UBAR \cite{yang2021ubar}, and PPTOD \cite{su-etal-2022-multi}, restructure all sub-tasks into a single sequence prediction through pre-training and fine-tuning. However, supervised fine-tuning (SFT) focuses more on learning at the token level than on the particular requirements, which limits the model's ability to complete specific tasks.

\paragraph{RL-Based Policy Learning.} RL can be leveraged to enhance model performance by tailoring it to the specific requirements of TOD tasks. However, RL models face challenges due to large action spaces and sparse rewards \cite{feng2023fantastic,zhang2019budgeted,wu2019switch}. Some studies use deep reinforcement learning (DRL) methods like Deep Q-Networks (DQN) \cite{peng2018deep,jang2021gpt} to improve policies with simulated user interactions. Hierarchical RL (HRL) breaks tasks into sub-tasks, creating a policy hierarchy \cite{peng-etal-2017-composite,liu2020gochat}, while feudal RL (FRL) abstracts state and action spaces for more general policies \cite{gao2018neural,casanueva2018feudal}. These methods primarily focus on dialogue policy learning with complex algorithmic designs and often lack a robust understanding of user intentions, resulting in suboptimal performance.

\paragraph{Reward Design for TOD.} Recent studies have found offline RL to be a promising method for stabilizing training with static datasets \cite{snell2022offline, feng2023fantastic}. Following the offline principle, many methods design rewards at the dialog and turn level when a goal is achieved \cite{kwan2023survey, lu2019goal, tang2018subgoal}, but reward signals remain sparse. Inverse reinforcement learning (IRL) and reward shaping techniques have been introduced to learn denser rewards and encourage faster learning \cite{li2020rethinking, takanobu2019guided}. However, IRL can be computationally intensive, and reward shaping might result in unintended behaviors if not carefully designed \cite{arora2021survey, gupta2024behavior}. Alternatively, some methods employ rewards for every token, which may lack semantic significance towards the dialogue goal \cite{yu2023krls, gupta2024behavior}. Our approach provides progressive rewards directly towards the dialogue goal.

\paragraph{Large Language Models for TOD.} LLMs have demonstrated impressive capabilities in understanding and generating text for various tasks \cite{ouyang2022training,DBLP:journals/corr/abs-2303-08774, chowdhery2023palm,wu2024perhaps}. 
However, LLMs underperform compared to specialized task-specific models \cite{hudevcek2023large, NEURIPS2023_c5601d99,wu-etal-2024-lamini}. Fine-tuning LLMs for specific tasks is also computationally inefficient. All these reasons lead to a growing interest in prompt engineering approaches that leverage in-context learning without requiring parameter updates \cite{wei2022chain, wang2022self, yao2024tree,DBLP:journals/corr/abs-2401-06468}. Yet, LLMs still tend to perform less effectively \cite{yang2024harnessing}. 

\section{Preliminary}

\subsection{Supervised Fine-Tuning for TOD}
The TOD task is typically modeled as an E2E problem and addressed by a seq2seq model (e.g.\ T5) using supervised fine-tuning (SFT). The input of the model can be represented as $\text{I}_t = [\text{prefix}:u_{t-1}:bs_{t-1}:da_{t-1}:sr_{t-1}:u_t]$, where \( [\cdot:\cdot] \) denotes the concatenation operator, \( u_t \) represents the current user utterance, \( bs_{t-1} \), \( da_{t-1} \), and \( sr_{t-1} \) represent the belief state (BS), dialogue act (DA), and system response (SR) at turn \( t-1 \) respectively. The prefix instruction is ``translate dialogue to belief state, dialogue action, and system response: [input]''. The model is fine-tuned to maximize the likelihood of successively generating correct BS, DA, and SR given the input:

\begin{equation}
\mathcal{L}_\theta = \sum_{t=1}^{T} \log P(bs_t, da_t, sr_t \mid \text{I}_t; \theta),
\end{equation}
where $\theta$ represents the parameters of the model. 

\subsection{Reinforcement Learning for TOD}
Formally, the RL approaches for TOD tasks operate within a Markov Decision Process (MDP) \cite{kaelbling1998planning} characterized by the tuple \(\langle S, A, P, R, \gamma \rangle\). The state space \( S \) can be represented as a set of states \(\mathbf{s}_i = \{ s_1, s_2, \ldots, s_k \}\), where each state includes the dialogue context and history up to the current time step. Each turn in the dialogue is considered an independent episode. An action \( a_{\Delta t} \in A \) is the \(\Delta t\)-th action taken during an episode, which corresponds to selecting the next token in the dialogue. Transition probability \( P(s' \mid s, a) \) is the probability of transitioning to state \( s' \) given action \( a \) and state \( s \). The discount factor \(\gamma \in [0, 1]\) is used to weigh future rewards. The SFT model is used to initialize a policy network \(\pi\), which is subsequently refined to maximize the reward \( R \), using algorithms such as proximal policy optimization (PPO) \cite{schulman2017proximal}.

\begin{figure*}[h]
    \centering
    \includegraphics[width=\textwidth]{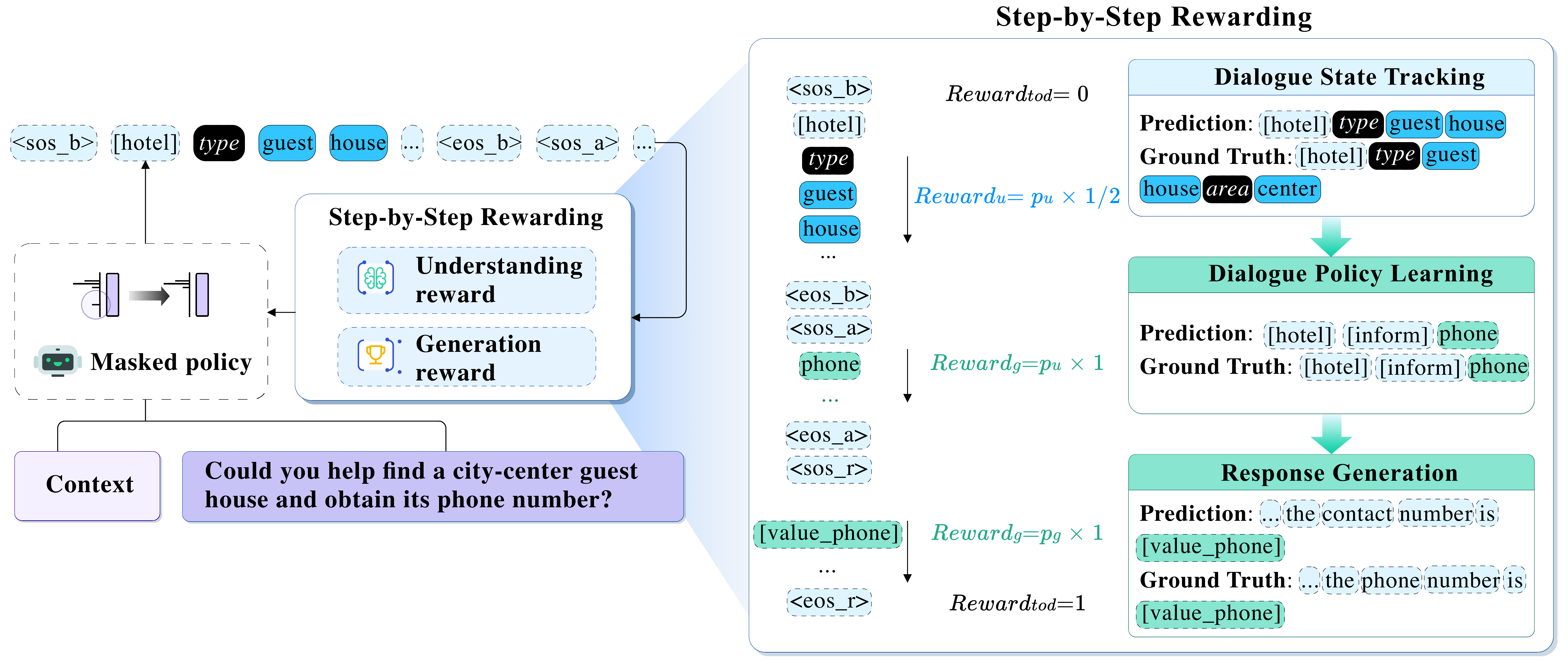}
    \caption{
        Overview of our approach. \textbf{Left}: We use the masked policy to optimize understanding and generation end-to-end with our reward function for TOD systems. \texttt{Context} is the concatenation of belief state (BS), dialogue act (DA), and system response (SR) at the previous turn. Special characters like \texttt{<sos\_$b$>}, \texttt{<sos\_$a$>}, and \texttt{<sos\_$r$>} denote the start of BS, DA, and SR, while \texttt{<eos\_$b$>}, \texttt{<eos\_$a$>}, and \texttt{<eos\_$r$>} denote their endings. \textbf{Right}: The designed reward function provides step-by-step rewards for understanding and generation tasks. $Reward_{u}$ refers to \autoref{eq:understanding}, $Reward_{g}$ refers to \autoref{eq:generation}, and $Reward_{tod}$ refers to \autoref{eq:tod}.
        }
    \label{fig:overview}
\end{figure*}

\section{Main Method}

We aim to enhance TOD systems using a combination of SFT and RL. While SFT can provide a stable initial base for RL \cite{ramamurthy2023reinforcement, yu2023krls, NEURIPS2023_c5601d99}, it equally treats every ground-truth token as an objective, without prioritizing task-specific goals. We utilize RL to refine the model to optimize for task completion.

In TOD tasks, accurately understanding user needs (i.e., belief states) is crucial for generating appropriate dialogue acts, which are essential for producing system responses that meet current needs and effectively drive the conversation forward. However, existing RL methods often focus solely on optimizing dialogue policy learning  \cite{NEURIPS2023_c5601d99, TakanobuLH20} or response generation \cite{yu2023krls}, neglecting the importance of understanding and the interdependence between understanding and generation. Moreover, these methods typically use sparse rewards at the dialogue or turn level \cite{kwan2023survey, lu2019goal, tang2018subgoal, DBLP:journals/corr/abs-2311-18232}. 

Task completion metrics evaluate whether the model correctly generates informable and requestable slot values defined in the dialogue schema, reflecting its performance in understanding and generation tasks. The policy model's sequence generation process involves continually satisfying these lists. Inspired by these metrics, we hypothesize that providing progressive task-oriented rewards during token generation for understanding and generation tasks can enhance TOD systems. The model architecture and our reward function are illustrated in \autoref{fig:overview}. In \autoref{sec:Metrics}, we explain how these metrics are measured to support our reward function design. In \autoref{sec:Reward}, we show how our reward function provides continuous, step-by-step feedback, guiding the E2E model through understanding and generation tasks for a more coherent and responsive dialogue system.


\subsection{Task Completion Metrics}
\label{sec:Metrics}
An \textit{informable} list and \textit{requestable} list are commonly predefined for dialog goals in datasets, such as In-Car and MultiWOZ. The \textit{informable} list contains slots and their values representing the user's requirements. For example, a user's preference for a restaurant is characterized by a ``cheap'' value for the ``price range'' slot. The \texttt{Inform} metric evaluates whether the system accurately learns user demands as defined in the \textit{informable} list and then provides a suitable entity in response. The \textit{requestable} list includes user-requested values, such as ``postcode''. The \texttt{Success} metric measures whether the generated DAs or SRs contain all attributes in the requestable list. Therefore, we believe that a slot-value-specific reward derived from the \textit{informable} list can enhance the system's understanding of user needs, while the value-specific reward based on the requestable list can improve responsiveness to user requests. Accordingly, we introduce the design of a progressive reward function combining the \emph{understanding} reward for DST, as well as the \emph{generation} reward for DPL and RG.

\subsection{Step-by-Step Goal-Oriented Reward} \label{sec:Reward}
\paragraph{Understanding Reward.}
We design the understanding reward for DST by measuring the growing proportion of correctly identified slot-value pairs in the \emph{informable} list during token (action) generation. This reward function directly reflects how well the system understands the user's needs, which is closely related to the goals of DST. Formally, we denote \( SV_{gt} \) as the set of ground-truth slot values in the current turn and \( \hat{SV} \) as the set of predicted ones during the token generation: 

\begin{equation}
R_{u} = \frac{|SV_{gt} \cap \hat{SV}| \cdot \rho_{u}}{|SV_{gt}|},
\label{eq:understanding}
\end{equation}
where 
$\rho_{u}=\exp \left(-\alpha \cdot \frac{|SV_{gt}\setminus\hat{SV}|}{|SV_{gt}|}\right)$ 
represents a penalty based on the discrepancy between the number of predicted slot-value pairs and ground truth slot-value pairs, with \( \alpha \) being a tunable parameter that controls the sensitivity of this penalty. The function provides a dense reward that progressively reflects the accuracy of DST.

\paragraph{Generation Reward.}
We observe that the accuracy of both DPL and RG depends on how many values in their generations are correctly included in the \emph{requestable} list. Therefore, we set the same reward function for these two generation tasks. The reward for DPL and RG is the increasing inclusion of values in the user \emph{requestable} list during each token generation, which measures the system's ability to fulfill user requests continuously. Formally, \( S_{gt} \) is all ground-truth user request values in the current turn, and \( \hat{S} \) denotes the predicted values during token generation:

\begin{equation}
R_{g} = \frac{|S_{gt} \cap \hat{S}|\cdot \rho_{g}}{|S_{gt}|},
\label{eq:generation}
\end{equation}
where the penalty term \( \rho_{g}=\exp \left(-\beta \cdot \frac{|S_{gt}\setminus\hat{S}|}{|S_{gt}|}\right) \) peralizes the difference between the number of generated values and values in the ground-truth requestable list, and \( \beta \) is a tunable parameter that controls the sensitivity of this penalty. The function provides a dense reward that progressively reflects how well the generation completes.

\paragraph{TOD Reward.}
To offer a comprehensive reward that evaluates both the understanding and generation performance, we define the TOD reward as a weighted combination of the understanding reward $R_{u}$ and the generation reward $R_{g}$: 
\begin{equation}
R_{tod} = \frac{|SV_{gt} \cap \hat{SV}| \cdot \rho_{u} + 
|S_{gt} \cap \hat{S}| \cdot \rho_{g}}{|SV_{gt}| + |S_{gt}|}.
\label{eq:tod}
\end{equation}

The combined reward function encourages balanced optimization of both the understanding (DST) and the generation (DPL, RG), which enhances the global robustness of TOD systems. The use of dense rewards derived from the informable and requestable lists ensures continuous feedback during token-level generation. Unlike sparse rewards that only provide feedback at the end of dialogues, our approach offers step-by-step rewards, accelerating the learning process. The progressive nature of the rewards, based on the discrepancies $\rho_{u}$ and $\rho_{g}$, helps make incremental improvements.

\paragraph{Reward Shaping.}
To prevent the policy network \( \pi \) from straying too far from the initial model \( \pi_{\text{o}} \), we also add a KL constraint to balance the reward. Formally, the final RL reward function is:
\begin{equation}
R_{total} = R_{t}- \beta D_{KL}({\pi} \parallel {\pi_{\text{o}}}),
\label{eq:bata}
\end{equation}
where \(\beta\) is dynamically adapted during training.


\paragraph{Optimization.}
We use natural language policy optimization (NLPO) \cite{ramamurthy2023reinforcement}, which is an extension of PPO. NLPO incorporates action elimination through a parameterized-masked approach. It learns to mask out less relevant tokens using top-p sampling, which restricts the token set to those with a cumulative probability above a specified threshold. NLPO maintains a separate masked policy that updates periodically, providing an additional constraint to ensure the selection of more task-relevant actions.

\begin{table*}[t]\small
\centering
\setlength{\tabcolsep}{5pt}
\begin{tabular}{lcccccccccccc}
\toprule
\multirow{2}{*}{\textbf{Method}} & \multicolumn{4}{c}{\textbf{MultiWOZ2.0}} & \multicolumn{4}{c}{\textbf{MultiWOZ2.1}} & \multicolumn{4}{c}{\textbf{In-Car}}\\
\cmidrule(lr){2-5} \cmidrule(lr){6-9} \cmidrule(lr){10-13}
   & Inform & Succ. & BLEU & Comb. & Inform & Succ. & BLEU & Comb. & Match & SuccF1 & BLEU &Comb.\\
\midrule
\multicolumn{9}{l}{\textup{\textbf{E2E}}} \\
SimpleTOD  & 84.4 & 70.1 & 15.0 & \phantom{0}92.3 & 85.0 & 70.5 & 15.2 & \phantom{0}93.0 & - & - & - & - \\
DoTS  & 86.6 & 74.1 & 15.1 & \phantom{0}95.5 & 86.7 & 74.2 & 15.9 & \phantom{0}96.3 & - & - & - & -\\
PPTOD  & 89.2 & 79.4 & 18.6 & 102.9 & 87.1 & 79.1 & 19.2 & 102.3 & - & - & - & -\\
UBAR$^{\dag}$  & 85.1& 71.0 & 16.2 & 94.3 & 86.2 & 70.3 & 16.5 & \phantom{0}94.7 & - & - & - & -\\
LABES  & - & - & - & - & 76.9 & 63.3 & 17.9 & \phantom{0}88.0 &  \underline{85.8} & 77.0 & 22.8 & 104.2 \\
SPACE-3$^{**}$ & 88.7 & 78.7 & 16.3 & 100.0 & 90.9 & 81.0 & 16.8 & 102.7 & 84.7 & 79.6 & 18.6 & 100.7\\
SPACE-3 & \underline{95.3} & 88.0 & \underline{19.3} & \underline{111.0} & \underline{95.6} & 86.1 & \underline{19.9} & \underline{110.8} & 85.3 & 83.2 & \underline{22.9} & 107.1\\
GALAXY$^*$  & 93.1 & 81.0 & 18.4 & 105.5 & 93.5 & 81.7 & 18.3 & 105.9 & 81.9 & 83.3 & 22.0 & 104.6\\
GALAXY  & 94.4 & 85.3 & \textbf{20.5} & 110.4 & 95.3 & 86.2 & \textbf{20.0} & \underline{110.8} & 85.3 & 83.6 & \textbf{23.0} & \underline{107.4}\\
\hdashline
\multicolumn{9}{l}{\textup{\textbf{RL}}} \\
MinTL   & 84.9 & 74.9 & 17.9 & \phantom{0}97.8 & - & - & - & - & - & - & - & - \\
GPT-Critic  & 90.1 & 76.6 & 17.8 & 101.1 & - & - & - & - & - & - & - & -\\
FanReward  & 93.1 & 83.9 & 18.0 & 106.5 & - & - & - & - & - & - & - & -\\
\hdashline
Ours$_{base}$  & 92.1 & \underline{88.3} & 16.6 & 106.9 & 92.7 & \underline{88.5} & 16.2 & 106.8 & 84.3 & \underline{83.8} &  22.8 & 106.9\\ 
Ours$_{large}$  & \textbf{96.1} & \textbf{92.4} & 17.2 & 
\textbf{111.5} & \textbf{96.9} & \textbf{91.1} & 16.9 & \textbf{110.9} & \textbf{86.2} & \textbf{86.1} & \textbf{23.0} & \textbf{109.2}\\ 

\bottomrule
\end{tabular}
 \caption{Performance comparison on MultiWOZ2.0, MultiWOZ2.1 and In-Car datasets. $\dag$: The results of UBAR are obtained using the models provided by the authors. $*$: \text{The results of GALAXY$^{*}$ are presented without pre-training.} $**$: \text{The results of SPACE-3$^{**}$ are results without pre-training, reimplemented using their public code.}}
\label{tab:main_table}
\end{table*}


\section{Experiments}

\subsection{Dataset}
\label{sec:dataset}
We conduct experiments on two popular task-oriented dialog benchmarks: MultiWOZ \cite{budzianowskietal2018multiwoz,conflrecEricGPSAGKGKH20}, and In-Car Assistant (In-Car) \cite{eric2017key}. The MultiWOZ datasets are a challenging benchmark for evaluating TOD systems with seven domains: attraction, hotel, hospital, police, restaurant, taxi, and train. The dataset is split into 8,438 dialogues for training, and 1,000 each for validation and testing. We use two versions, i.e.\ MultiWOZ2.0 and MultiWOZ2.1, to evaluate our model. The In-Car dataset comprises 3,031 multi-turn dialogues across three specific domains suitable for an in-car assistant: calendar scheduling, weather information retrieval, and point-of-interest navigation. The dialogues in In-Car are more natural and diverse. We split the dataset into training/validation/testing sets containing 2425/302/304 dialogs respectively as previous works do. Following the data preprocessing procedure from \cite{zhang2020task}, delexicalized responses are utilized in our work to help the model learn generalizable parameters.

\subsection{Evaluation Metrics}

In this work, we evaluate our models on MultiWOZ and In-Car benchmarks as described in \autoref{sec:dataset}.
For MultiWOZ, we report \texttt{Inform} and \texttt{Success} as introduced in \autoref{sec:Metrics}. Additionally, we report \texttt{BLEU} \cite{papineni2002bleu} that is used to measure the fluency of the generated response. Consequently, we report (\texttt{Comb}) that is computed by \texttt{(Inform + Success) ×0.5 + BLEU} as an overall quality measure. For In-Car, we leverage \texttt{Match} to measure if a system can track all correct states to satisfy the user. \texttt{SuccF1} improves on the \texttt{Success} by considering both how completely (recall) and accurately (precision) the system handles requests. 

Both \texttt{Inform} and \texttt{Match} evaluate the system's understanding of user requirements, but \texttt{Inform} further focuses on providing correct entities based on the understanding, while \texttt{Match} ensures accurate dialogue state tracking. Similarly, both \texttt{Success} and \texttt{SuccF1} assess the system's ability to fulfill user requests, but \texttt{Success} measures whether all user requests are met, whereas \texttt{SuccF1} balances precision and recall to gauge response accuracy and completeness. The design of our reward function aligns with all these task completion metrics for the dialogue datasets.


\subsection{Baselines}
We comprehensively evaluate our approach by comparing it with a wide array of methods on the MultiWOZ2.0 and MultiWOZ2.1 datasets. We select several prominent end-to-end (E2E) models as baselines, including SimpleTOD \cite{hosseini2020simple}, DoTS \cite{jeon2021domain}, PPTOD \cite{suetal2022multi}, UBAR \cite{yang2021ubar}, GALAXY \cite{he2022galaxy}, and SPACE-3 \cite{he2022unified}. Furthermore, we compare our approach with current representative reinforcement learning (RL) models, including MinTL \cite{linetal2020mintl}, GPT-Critic \cite{jang2021gpt}, and FanReward \cite{feng2023fantastic}. These approaches demonstrate the potential of RL for TOD models and offer valuable comparative perspectives. We compare our approach against several strong baselines on the In-Car dataset, including LABES \cite{zhangetal2020probabilistic}, SPACE-3 \cite{he2022unified}, and GALAXY \cite{he2022galaxy}. Additionally, we present results for GALAXY and SPACE-3 without pre-training across the three datasets to provide a comprehensive evaluation. Two pre-trained models, the Flan-T5-Base and Flan-T5-Large \cite{chung2024scaling}, are utilized as the backbone of our approach (see \autoref{sec: appendixA}).

\begin{figure*}[h]
    \centering
    \includegraphics[width=\textwidth]{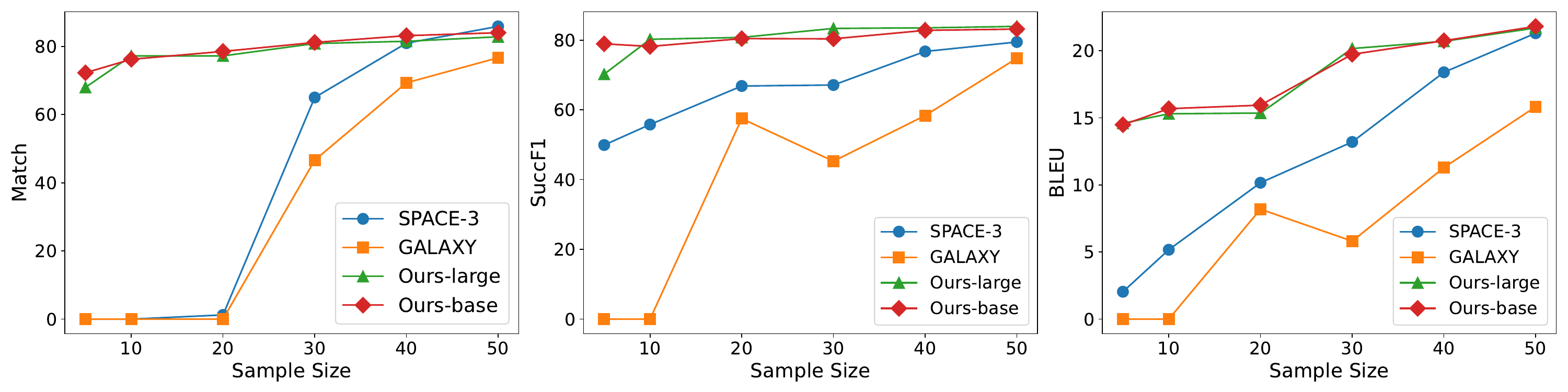}
    \caption{Results of low-resource experiments. 5\% (121 dialogues), 10\% (242 dialogues), 20\% (485 dialogues), 30\% (727 dialogues), 40\% (970 dialogues), and 50\% (1212 dialogues) of training data is used to train each model. Results are shown as mean values over five runs.}
    \label{fig:few_shot}
\end{figure*}

\subsection{Main Results}
As shown in \autoref{tab:main_table}, our approach achieves new state-of-the-art results across all datasets in the combined score (\texttt{Comb}). These gains are primarily due to the increased \texttt{Inform} and \texttt{Success} rates. We achieve competitive \texttt{BLEU} scores on MultiWOZ, possibly because our approach focuses on understanding and responsiveness over fluency. However, our model performs best on the In-Car dataset, demonstrating its ability to generate fluent responses. Additionally, our model is compared with two strong baselines, SPACE-3 and GALAXY, both of which utilize multiple TOD datasets for pre-training and are subsequently fine-tuned on the MultiWOZ and In-Car datasets. Our approach, without a pre-training step, demonstrates a significantly greater improvement compared to their results without pre-training. For instance, on the MultiWOZ2.0 dataset, our method achieves an increase of +3 points in the \texttt{Inform} rate and +11.4 points in the \texttt{Success} rate. This indicates that our approach, which offers step-by-step rewards for RL, effectively helps boost TOD task completion.


\subsection{Ablation Study}
We conduct an ablation study on the MultiWOZ2.0 dataset to evaluate the effectiveness of our progressive goal-oriented reward mechanism. As shown in \autoref{tab:ablation}, when the understanding reward \(R_{u}\) is removed, leaving only the generation reward, the combined score drops significantly to \(105.2\). This substantial decrease highlights the crucial role of immediate feedback during dialogue state tracking. When we remove the generation reward \(R_{g}\), the combined score decreases by \(6.1\) points. This suggests that the generation reward is also important for task completion. If \(R_{u}\) and \(R_{g}\) are all removed, that is the result of only SFT, the performance degrades to a lower value. Overall, the study demonstrates that our progressive reward mechanism greatly improves the system's ability to perform understanding and generation tasks.



\section{Analysis and Discussion}

In this section, we first explore how well our model performs in low-resource settings (\autoref{sec:low_resource}). Besides, we show that our approach can be integrated into recent state-of-the-art LLMs for better performance (\autoref{sec:integrate}). Lastly, we conduct human evaluation on the generated outputs (\autoref{sec:human_eval}).

\subsection{Low-Resource Evaluation}
\label{sec:low_resource}
Due to the challenge of creating extensive, well-annotated dialogue datasets for real-world applications, we also explore the performance of our approach with limited training samples. We train models using the In-Car dataset and randomly sample 5\%, 10\%, 20\%, 30\%, 40\%, and 50\% of the training data. Our approach is benchmarked against two robust baselines, SPACE-3 and GALAXY. Both models are initialized with their pre-trained versions and subsequently fine-tuned using the sampled datasets from the In-Car dataset. Our model is trained through SFT and RL stages both with the sampled data. To ensure fairness, all models are trained for 30 epochs. \autoref{fig:few_shot} presents the experimental results. As illustrated, our approach consistently outperforms the baselines across all sample sizes on the metrics of \texttt{Match}, \texttt{SuccF1}, and \texttt{BLEU}. The performance advantage is especially prominent when the training data is limited. This suggests that our model demonstrates enhanced generalizability and is more apt for tackling new \texttt{TOD} tasks. It is noted that results between \texttt{Ours-large} and \texttt{Ours-base} are similar. This may be because, in a low-data setting, the smaller model (\texttt{Ours-base}) better utilizes the available data. In contrast, the larger model (\texttt{Ours-large}) may not have been fully trained with the limited data, leading to no significant performance improvement.

\begin{table}[t]
\centering
\begin{tabular}{lcccc}
\toprule
 Model  & Inform & Succ. & BLEU & Comb. \\
\midrule
Ours  & 96.1 & 92.4 & 17.2 & 111.5  \\
 $-R_{u}$ & 91.2 & 87.0 & 16.1 & 105.2 \\
 $-R_{g}$ & 92.1 & 87.5 & 15.6 & 105.4 \\
 $-R_{u}-R_{g}$  & 86.0& 81.8 & 17.2 & 101.1 \\
\bottomrule
\end{tabular}
 \caption{Ablation results on MultiWOZ2.0.}
\label{tab:ablation}
\end{table}


\subsection{Integration with LLMs}
\label{sec:integrate}
Recently, LLMs have led to remarkable advancements in NLP, demonstrating impressive emergent abilities. However, LLMs often underperform compared to specialized models for TOD tasks \cite{hudevcek2023large, NEURIPS2023_c5601d99}. We utilize few-shot dialogue examples of the training set of MultiWOZ2.0 to prompt LLMs. As shown in the top half of \autoref{tab:llm}, the powerful representatives of LLMs, including Codex (code-davinci-002) \cite{abs210703374}, ChatGPT (gpt-3.5-turbo) \cite{Ouyang0JAWMZASR22}, Claude (Claude 3 sonnet) \cite{anthropic2024claude}, and GPT-4o\footnote{\url{https://openai.com/index/hello-gpt-4o/}}, do not perform as well as our model. Additionally, fine-tuning LLMs for TOD systems can also be resource-intensive and computationally inefficient. There has been a surge of research interest that combines LLMs with small models for specific applications. We follow the recent work DSP \cite{ NEURIPS2023_c5601d99}, which utilizes a small tunable model for dialogue policy learning to generate dialogue acts. The dialogue acts are used as hints to prompt LLMs to generate the final response. To demonstrate the advantages of our approach, we employ our generation reward function to enhance the policy learning for the small model. To be fair, we use the same data setup as the DSP method -10\% of the training data for SFT and RL training. Our approach shows superior performance on the MultiWOZ2.0 dataset. The \texttt{Success} rate is 8.9\% higher than the DSP results reported in their work. This demonstrates that our method has superior generation capabilities, leading to more effective task completion.

\begin{table}[t]
\centering
\small
\setlength{\tabcolsep}{5pt}
\begin{tabular}{lcccc}
\toprule
\multirow{2}{*}{\textbf{Method}} & \multicolumn{4}{c}{\textbf{MultiWOZ2.0}}\\
\cmidrule(lr){2-5}
   & Inform & Succ. & BLEU & Comb.  \\
\midrule
Codex$^{\dag}$ & 76.7 & 41.5 & \phantom{0}7.7 & 66.8  \\
ChatGPT$^{\dag}$   & 71.8 & 44.1 & \phantom{0}10.5 & 68.4  \\
Claude  & 78.3 & 41.2 & \phantom{0}2.9 & 62.7  \\
GPT-4o   & 77.0 & 53.1 & \phantom{0}5.2 & 70.3  \\
\hdashline
DSP w/ ChatGPT$^{\dag}$ & \underline{95.3} & 82.3 & \underline{10.9} & 99.6  \\
Ours w/ ChatGPT   & 95.1 & \underline{91.2} & \phantom{0}9.8 & \underline{102.9} \\
\hdashline
Ours$_{large}$  & \textbf{96.1} & \textbf{92.4} & \textbf{17.2} & \textbf{111.5} \\
\bottomrule
\end{tabular}
 \caption{Performance comparison on MultiWOZ2.0 based on different LLMs. $^{\dag}$: The results are reported in the work of DSP\cite{NEURIPS2023_c5601d99}. }
\label{tab:llm}
\end{table}

\begin{figure}[h]
    \centering
    \includegraphics[width=0.5\textwidth]{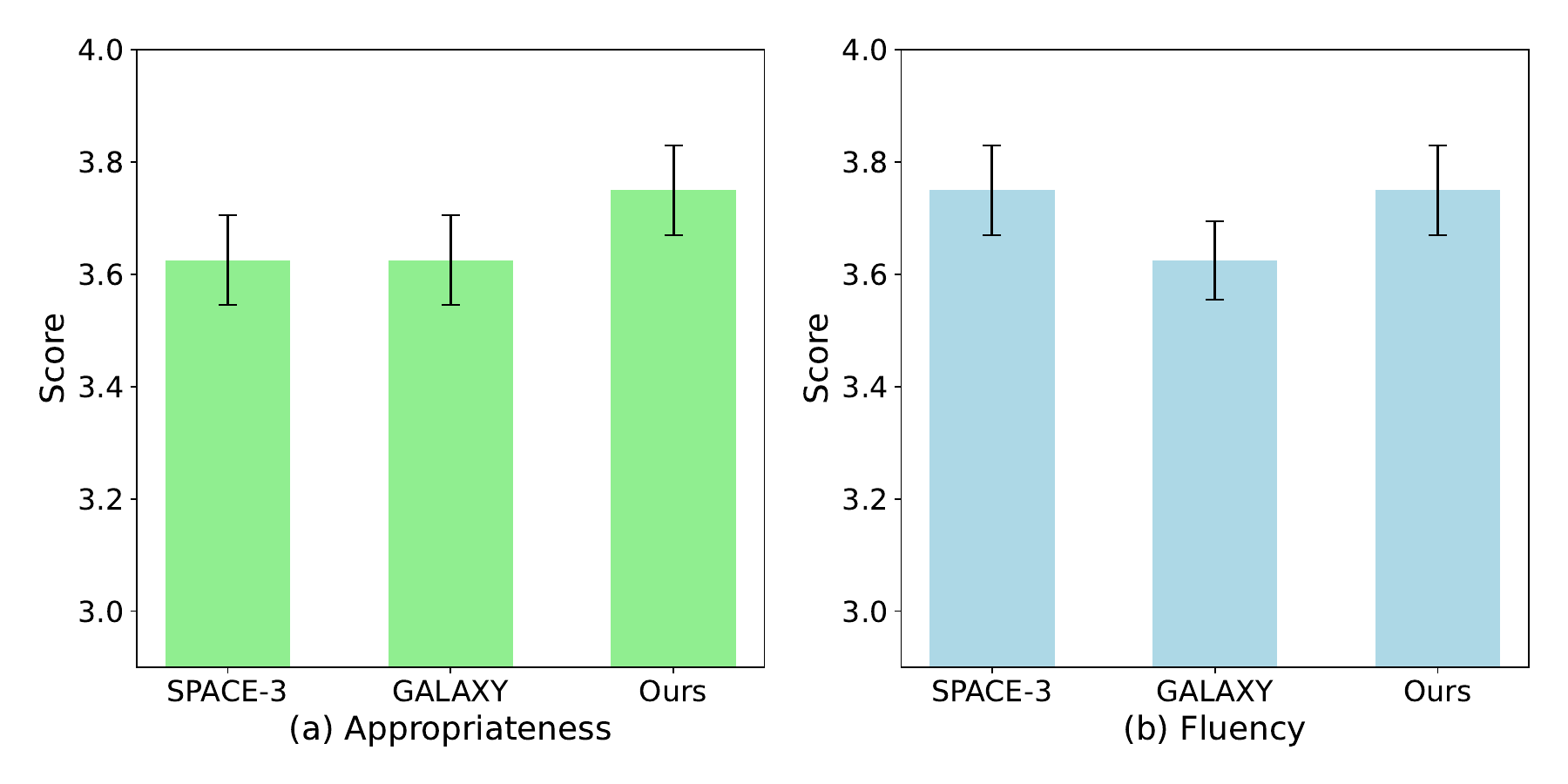}
    \caption{The human evaluation results regarding appropriateness and fluency. The numbers represent the average and the standard deviation for each method.}
    \label{fig:human_evaluation}
\end{figure}


\subsection{Human Evaluation}
\label{sec:human_eval}
The automatic evaluation metric like \texttt{BLEU} might not be able to accurately evaluate the generation quality \citep{freitag-etal-2021-experts,freitag-etal-2022-results,DBLP:journals/corr/abs-2307-03025}. To thoroughly evaluate our approach, we conduct a human evaluation of the generated responses using our developed platform. We compare our model with the top two baselines, SPACE-3 and GALAXY, from \autoref{tab:main_table} on the In-Car dataset. Following previous works \cite{zhang2020task,ramachandranetal2022caspi,jang2022gptcritic}, we use two metrics: 1) Appropriateness: how well the response fits the dialogue context, and 2) Fluency: the clarity and coherence of the response. We randomly select 50 dialogue turns from the test set, showing each turn and its history to 6 evaluators. Evaluators score each response on a 5-point Likert Scale (1 to 5), where 1 represents the lowest quality and 5 represents the highest quality. Importantly, the evaluators are unaware of the model identities to ensure unbiased judgments. As shown in \autoref{fig:human_evaluation}, our model outperforms the baselines in terms of appropriateness and matches SPACE-3 in fluency. This aligns with the results in \autoref{tab:main_table} and further validates that our approach can not only achieve superior dialogue-task completion performance but also ensures high-quality responses.



\section{Conclusion}
We introduce a new approach for incorporating RL into TOD systems. Our approach focuses on improving both understanding and generation tasks by addressing challenges related to sparse and delayed rewards. We devise a progressive reward mechanism that combines understanding and generation rewards at the token level, facilitating gradual learning. Through extensive experiments on standard benchmarks using Flan-T5-Base and Flan-T5-Large backbones, we demonstrate the effectiveness of our approach and achieve state-of-the-art results on three widely used datasets.


\section{Limitations}
While our proposed approach using step-by-step rewards has shown promising results, it may struggle to capture all the nuances of TOD tasks fully. As a result, biases could be unintentionally introduced, causing the model to learn suboptimal strategies. In the future, it would be beneficial to develop a comprehensive reward model grounded in our reward function. Such a model can learn intricate patterns and enhance flexibility and adaptability.

Moreover,  the reward design in our approach relies on predefined \emph{informable} and \emph{requestable} lists in the dialogue schema. While this is a common practice in task-oriented dialogue systems, it is limited when extending to open-domain dialogues. Open-domain dialogues typically lack fixed slots and values, which makes it challenging to apply this reward mechanism effectively. In the future, it would be valuable to have a more generalizable approach that supports both task-oriented and open-domain dialogues in conversational agents.

\bibliography{latex/emnlp2024}

\newpage
\appendix

\label{sec:appendix}
\section{Implementation Details}
\label{sec: appendixA}
\paragraph{SFT Details.} In the supervised fine-tuning stage, we use a train batch size of 8 and an evaluation batch size of 16. We set the learning rate to \(2 \times 10^{-5}\) and train the model for a total of 30 epochs. For generation settings of inference, the maximum length is set to 256 tokens for the MultiWOZ dataset and 168 tokens for the In-Car dataset respectively. The shorter maximum length for the In-Car dataset is due to the relatively shorter utterances compared to those in the MultiWOZ datasets. It is important to note that the In-Car dataset does not include dialogue act annotations. Hence, we predict only the belief state and system response, formatting the input as $\text{I}_t = [\text{prefix}:u_{t-1}:bs_{t-1}:sr_{t-1}:u_t]$.


\paragraph{RL Details.} The policy network is trained for 20k episodes, with 5 epochs per batch for enforcement learning. The batch size is set to 8, and the learning rate is \(2 \times 10^{-6}\). We employ sampling with a top-k value of 50 during training. Following \cite{DBLP:journals/corr/abs-1909-08593}, the KL coefficient $\beta$ in \autoref{eq:bata} is dynamically adapt during training:

\begin{equation}
    e_t = \text{clip} \left( \frac{D_{KL}(\pi \| \pi_{o}) - \text{KL}_{\text{t}}}{\text{KL}_{\text{t}}}, -0.2, 0.2 \right),
\end{equation}

\begin{equation}
    \beta_{t+1} = \beta_t (1 + K_{\beta} e_t),
\end{equation}
where $\text{KL}_{\text{t}}$ is the KL divergence between initial model $\pi_{o}$ and current policy $\pi$. $\beta$ are initially set to 0.01. $K_{\beta}$ is the update rate which we set to 0.2 in our experiments.



\paragraph{Model and Implementation Details.}
We use Flan-T5 base ($\sim$250M parameters) and Flan-T5 large ($\sim$780M parameters) models as the backbone, which are the extensions of the T5 model designed to enhance performance on a wide range of natural language processing tasks. Our experiments are all run on a server equipped with 8 NVIDIA A800. 

\section{Reward Curve}
\label{sec: appendixB}

\begin{figure*}[h]
    \centering
    \includegraphics[width=\textwidth]{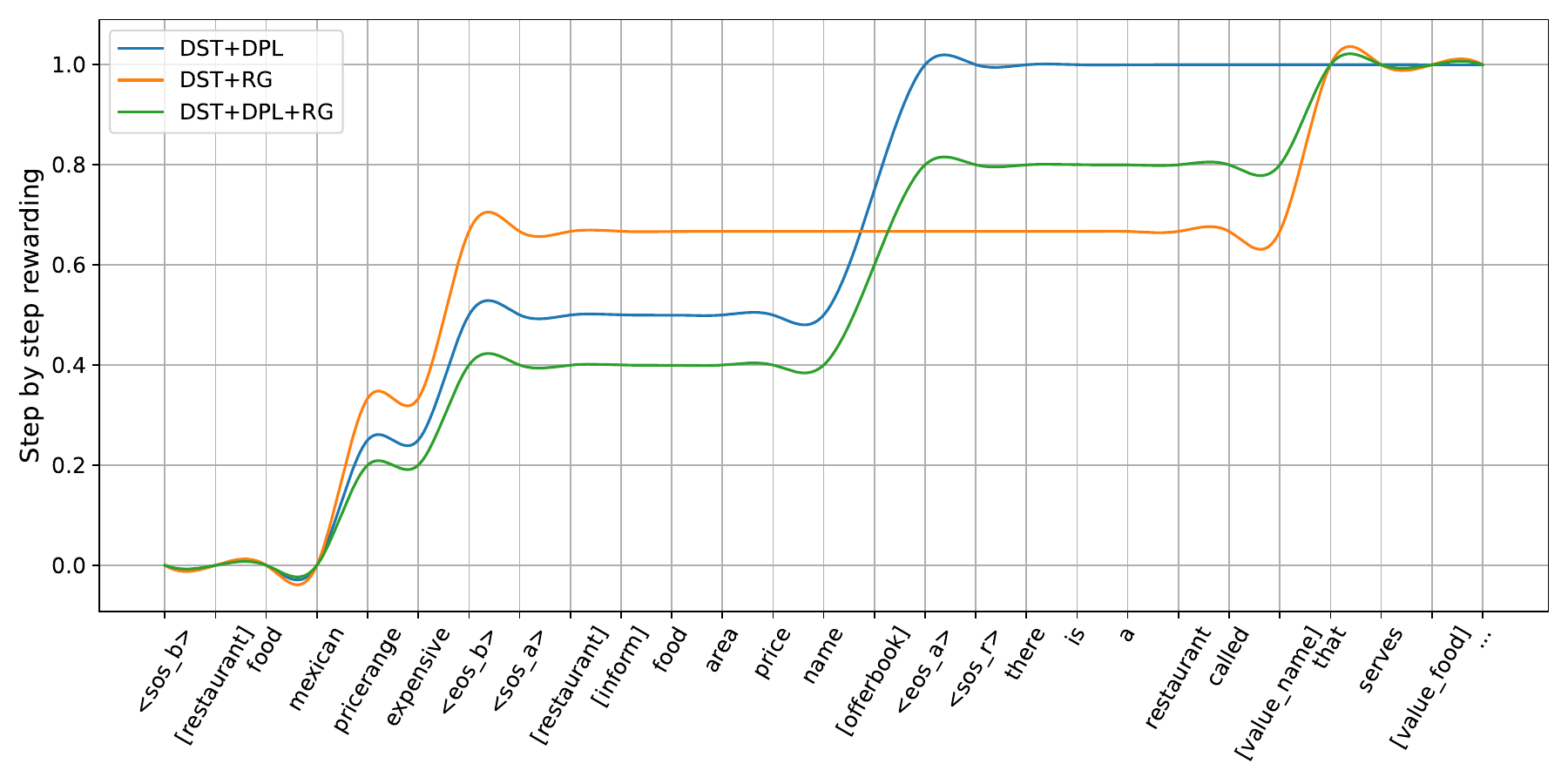}
    \caption{Reward accumulation for different tasks: DST+DPL, DST+RG, and DST+DPL+RG during token generation. Our reward function progressively provides important feedback for understanding and generation tasks.}
    \label{fig:reward_analysis}
\end{figure*}

\autoref{fig:reward_analysis} shows how rewards increase incrementally during token-by-token generation when completing slot-values or values. The plateau phases represent the process of generating a complete slot value or value. We present the reward patterns for three tasks: DSP, DPL, and RG. The curves demonstrate that our approach provides gradually increasing dense rewards for end-to-end models, effectively supporting understanding and generation tasks.






\section{Case Study}
\label{sec: appendixC}

\begin{figure*}[h]
    \centering
    \includegraphics[width=\textwidth]{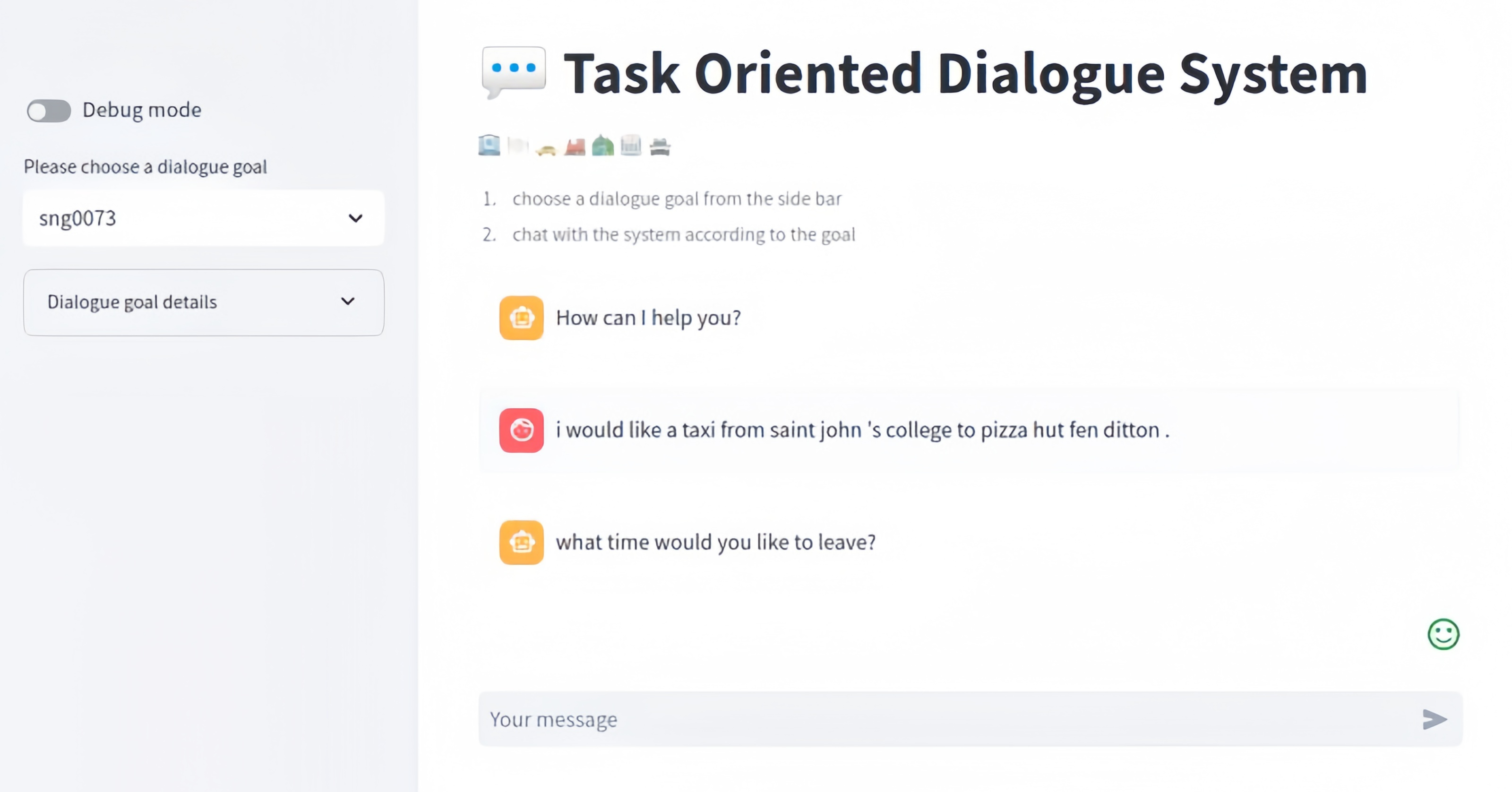}
    \caption{Our user interface allows users to evaluate the dialogue system. They can select dialogue goals, interact with the system, and provide feedback on the responses.}
    \label{fig:streamlit_interface}
\end{figure*}

To evaluate the effectiveness of our dialogue system, we develop a user interface using the Streamlit\footnote{\url{https://streamlit.io/}} as shown in \autoref{fig:streamlit_interface}. The interface allows users to select a dialogue goal and interact with the system according to that goal. Users assess the system's responses utilizing the evaluation methodology detailed in \autoref{sec:human_eval}.

\begin{figure*}[h]
    \centering
    \includegraphics[width=\textwidth]{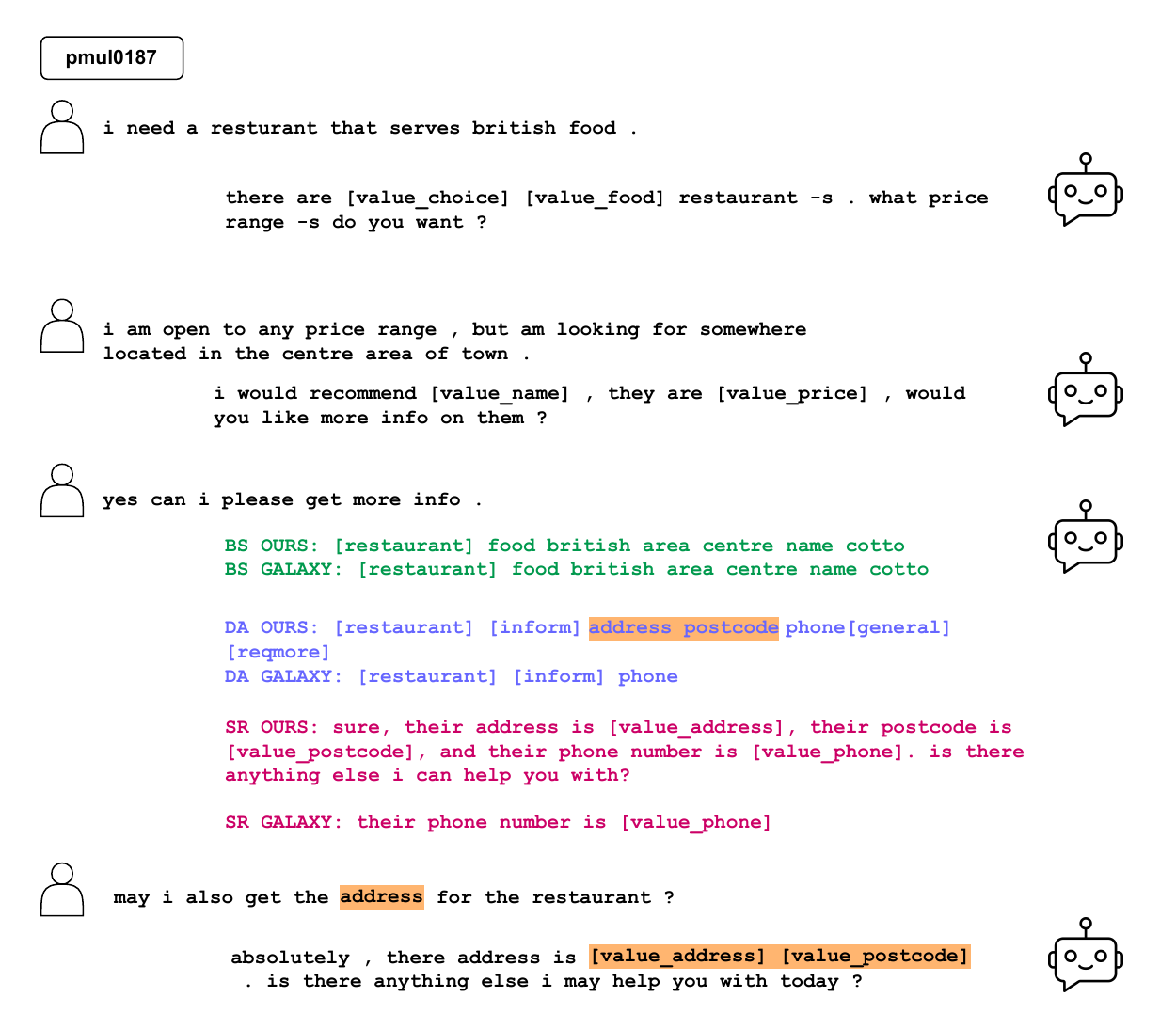}
    \caption{An example showing our model's effectiveness in predicting comprehensive restaurant information. Green, purple, and red text represent predicted results of our model and GALAXY for BS, DA, and SR respectively. }
    \label{fig:exmaple1}
\end{figure*}

We provide an example comparison from our model and GALAXY in \autoref{fig:exmaple1}. It illustrates a scenario in which our model generates more accurate and comprehensive results compared to GALAXY.


\section{Error Examples}
\label{sec: appendixD}

\begin{figure*}[h]
    \centering
    \includegraphics[width=\textwidth]{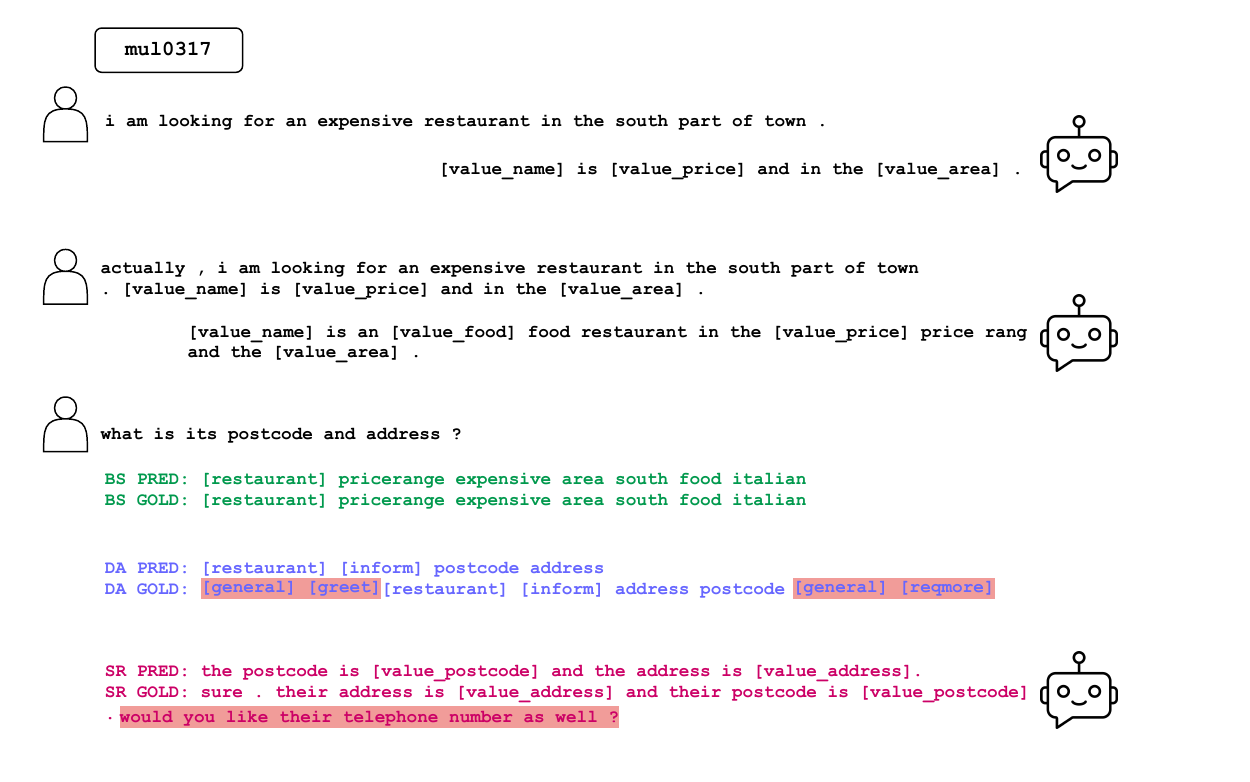}
    \caption{Error Example. Black text is the input context. Green, purple, and red text represent predicted (PRED) and ground truth (GOLD) for BS, DA, and SR. Red highlights indicate incorrect or missing key tokens.}
    \label{fig:error3}
\end{figure*}

We present a representative error example in our predicted results in \autoref{fig:error3}. We observe that the response of our model includes all the necessary value information for the task, but it lacks conversational fluency. This indicates that our designed reward function prioritizes task completion efficiency over dialogue naturalness. Future work could explore integrating metrics like BLEU into the reward function to enhance both task completion and conversational fluency.

\end{document}